\documentclass{article}

\usepackage{iclr2027_conference,times}

\usepackage[utf8]{inputenc} 
\usepackage[T1]{fontenc}    
\usepackage{hyperref}       
\usepackage{url}            
\usepackage{booktabs}       
\usepackage{amsfonts}       
\usepackage{nicefrac}       
\usepackage{microtype}      
\usepackage{xcolor}         
\usepackage{graphicx}
\usepackage{afterpage}

\title{CAPT: A Multi-task Continuous Autoregressive Transformer enabling Cross-dataset and Cross-species Transfer for Calcium Population Dynamics}

\author{%
  Xinhong Xu\thanks{Equal contribution.} \\
  Department of Automation\\
  Tsinghua University\\
  Beijing, China\\
  \And
  Yimeng Zhang\footnotemark[1] \\
  School of Life Sciences\\
  Tsinghua University\\
  Beijing, China
  \And
  Yuanlong Zhang\thanks{Corresponding author: \texttt{yuanlongzhang@tsinghua.edu.cn}} \\
  School of Life Sciences\\
  Tsinghua University\\
  Beijing, China
}

\iclrfinalcopy 
\begin{document}

\maketitle

\fancyhead{}
\renewcommand{\headrulewidth}{0pt}

\begin{abstract}

Calcium imaging has enabled large-scale recordings of neural population activity across diverse animals, laboratories, and experimental paradigms. However, these datasets remain highly fragmented: Existing approaches are often designed for individual datasets or tasks, making it unclear whether their learned representations are reusable for new calcium imaging trace datasets. To tackle this gap, we present \textbf{CAPT}, a \textbf{C}ontinuous \textbf{A}utoregressive \textbf{P}opulation \textbf{T}ransformer for calcium population dynamics. CAPT models continuous calcium traces directly through a continuous patch tokenization strategy and is trained autoregressively with a mean squared error objective, enabling end-to-end training and adaptation to diverse downstream tasks. We first pretrain CAPT on a large-scale mouse calcium imaging dataset and evaluate its transferability across independent mouse, larval zebrafish, and \textit{C. elegans} datasets collected by different laboratories. In these transfer settings, the pretrained backbone is frozen and only adaptation modules, i.e., neuron and session embeddings or task-specific heads, are updated. Across neural population forecasting and behavior decoding tasks, CAPT consistently outperforms specialized and general-purpose baselines, demonstrating that a single continuous autoregressive backbone can learn representations that transfer beyond the original pretraining distribution. Beyond predictive performance, multimodal analyses using NeuroPAL annotations in \textit{C. elegans} datasets show that CAPT embeddings form a shared functional space across datasets and capture cell-identity-related structure. Together, these results suggest that continuous autoregressive modeling provides an effective framework for learning reusable calcium population representations that transfer across datasets, experimental paradigms, and species. 

\end{abstract}

\section{Introduction}

Recent advances in neural recording technologies \citep{jun2017fully, steinmetz2021neuropixels, ye2025ultra, stringer2019spontaneous}, especially large-scale calcium imaging \citep{barson2020simultaneous, sofroniew2016large}, 
have made it possible to observe neural population activity across many neurons, animals, brain regions, and model organisms \citep{zong2022large,kim2017pan,musall2019single}. These recordings are rapidly accumulating across laboratories and species, including mice, larval zebrafish, and \textit{C. elegans}. As the scale and diversity of calcium imaging data continue to grow, an important challenge is how to develop models whose learned representations can be reused across recordings rather than being restricted to individual datasets or experimental settings. Recent neural population models have begun to exploit large-scale or multi-session recordings to learn shared representations for tasks such as neural forecasting and behavior decoding \citep{duan2025poco,azabou2025multi,willeke2026omnimouse}. 
However, the extent to which such learned representations remain useful when transferred to independently collected calcium imaging datasets is still poorly understood.

This challenge is particularly pronounced for calcium imaging because datasets can differ substantially in species, brain regions, imaging conditions, sampling rates, neuron counts, and experimental paradigms  \citep{grienberger2022twophoton,ahrens2013whole,kim2017pan}. A model trained on one dataset may therefore face pronounced distribution shifts when applied to another due to this heterogeneity. While recent neural models have shown encouraging results within specific datasets, organisms, or task families, it remains less studied whether a single frozen pretrained backbone can be reused across independently collected calcium imaging datasets while supporting multiple downstream tasks, such as neural population forecasting and behavior decoding \citep{azabou2025multi}. In particular, cross-species transferability remains largely underexplored in the current literature, leaving open how broadly representations learned from one neural population can remain useful across distinct experimental paradigms and biological systems.

To address this challenge, we propose \textbf{CAPT} (\textbf{C}ontinuous \textbf{A}utoregressive \textbf{P}opulation \textbf{T}ransformer), a continuous autoregressive model for calcium imaging population dynamics. \textbf{CAPT directly models continuous neural activity through a continuous patch tokenization strategy and is trained in an end-to-end, autoregressive manner with a mean squared error (MSE) objective, and is capable of performing diverse tasks.} We first pretrain CAPT on a large-scale mouse calcium imaging dataset. To systematically evaluate the transferability, we test CAPT on eight independent datasets beyond the pretraining data: three mouse datasets, three larval zebrafish datasets, and two \textit{C. elegans} datasets (Table \ref{tab:dataset_summary}, Appendix \ref{app:dataset}). In all transfer experiments, the pretrained backbone is frozen, and only adaptation components are updated: neuron and session embeddings for neural population forecasting and task-specific heads for behavior decoding. Following the aforementioned steps, we report that \textbf{CAPT consistently outperforms specialized and general-purpose baselines, including POYO+, POCO, and CalM, across both pretraining and transfer datasets}, and shows significantly better performance in most cases (Table \ref{tab:statistical_test}), while POYO+ and CalM are less directly applicable or degrade substantially under cross-species transfer. As a long-term goal of neural representation learning is to provide tools that facilitate biological discovery beyond prediction, we further test whether CAPT can support multimodal biological analysis by linking functional neural embeddings with NeuroPAL annotations as cell-identity labels in the \textit{C. elegans} datasets \citep{yemini2021neuropal}. Interestingly, beyond forecasting and decoding performance, we find that \textbf{CAPT embeddings recover identity-related structure and support cross-dataset neuron identity classification using only a linear support vector machine (SVM)} \citep{cortes1995support}\textbf{, suggesting that the model learns a shared functional embedding space across datasets and captures anatomical information.} By comparison, POCO embeddings show much weaker identity-related structure, with performance close to chance in the most challenging cross-dataset setting. These results indicate that CAPT not only improves predictive performance, but also provides embeddings that support biologically interpretable multimodal analysis.

In summary, our contributions are as follows:

\noindent\textbf{Continuous autoregressive modeling for calcium traces.}
We introduce CAPT, a continuous autoregressive transformer for calcium population dynamics with continuous patch tokenization and neuro-aware QKV modulation, and show that it achieves superior performance on diverse neural tasks and datasets against strong specialized and general-purpose baselines.

\noindent\textbf{Cross-dataset and cross-species transfer.}
We systematically evaluate the transferability of CAPT on multiple calcium imaging datasets spanning mice, larval zebrafish, and \textit{C. elegans}. With the pretrained backbone frozen, CAPT generalizes successfully across datasets and species.

\noindent\textbf{Biologically meaningful functional embeddings.}
We show that CAPT embeddings support multimodal analysis with NeuroPAL annotations in \textit{C. elegans}, forming a shared functional embedding space that captures anatomical cell-identity-related structure across datasets.

\vspace{-0.3cm} 
\section{Related Work}
\vspace{-0.1cm} 

\subsection{Neural dynamics and neuro-behavior analysis}
\vspace{-0.1cm} 

A long line of classical work has modeled neural population activities to characterize neural dynamics and their relationships to behavior. Latent dynamical systems and state-space models have been used to uncover low-dimensional structure for neural forecasting, trajectory reconstruction, and trial-level interpretation \citep{macke2011empirical, paninski2010new, churchland2012neural}. Gaussian-process-based methods extract smooth single-trial neural trajectories from population activity \citep{yu2008gaussian}. In parallel, neural decoding and encoding models relate population activity to behavioral variables, stimuli, choices, or movement kinematics \citep{wu2006bayesian, paninski2007statistical, pillow2008spatio}, providing important tools for studying neural representations and behavior. However, these models are often developed and evaluated within individual sessions, animals, or experimental settings.

\vspace{-0.3cm} 
\subsection{Transformer for neural data and neural foundation model}
\vspace{-0.1cm}

Recent work has begun to leverage Transformer architectures for large-scale representation modeling of invasive neural recordings. For spiking and electrophysiology data, NDT first showed that transformer architectures can model binned neural spiking activity with self-supervised sequence objectives \citep{ye2021representation}. This direction was extended by NDT2 for multi-context pretraining across spiking datasets \citep{ye2023neural}, by POYO for scalable neural population decoding with spike-level tokenization \citep{azabou2023unified}, and by Neuroformer for multimodal and multitask generative pretraining on spiking data \citep{antoniades2023neuroformer}. More recent models such as MtM and NEDS use multi-task masking to relate neural activity and behavior in large-scale Neuropixels recordings \citep{zhang2024towards, zhang2025neural}, while NeuroPaint studies neural inpainting by inferring unrecorded brain-area dynamics from multi-animal electrophysiology datasets \citep{xia2025inpainting}. These studies demonstrate the promise of scalable pretraining for neural data, but they are primarily developed for spiking or electrophysiology recordings, rather than continuous calcium traces.

For calcium imaging, POCO studies scalable neural forecasting through population conditioning and trains across calcium imaging datasets spanning mice, zebrafish, and \textit{C. elegans} \citep{duan2025poco}. POYO+ extends the POYO framework to multi-session decoding in large-scale mouse two-photon calcium imaging datasets \citep{azabou2025multi}. POYO-CAP explores calcium pretraining with cell-pattern-aware neuron selection for visual decoding \citep{bae2025decoding}. CalM introduces a self-supervised model based on vector quantization and a dual-axis transformer, enabling both forecasting and behavior decoding on calcium population traces \citep{xu2026self}, while OmniMouse studies multi-modal, multi-task calcium modeling at scale in mouse visual cortex data with deconvoluted trace \citep{willeke2026omnimouse}. Together, these works demonstrate the growing potential of large-scale pretraining and reusable representation learning for calcium population activity.

\vspace{-0.3cm} 
\subsection{Cross-dataset and cross-species neural modeling}
\vspace{-0.1cm}

An important question in neural representation learning is whether representations learned from one collection of recordings can remain useful across different datasets, animals, and experimental settings. In electrophysiology, MtM and NEDS demonstrate held-out-session or held-out-animal evaluation within standardized Neuropixels datasets \citep{zhang2024towards, zhang2025neural}. POSSM further demonstrates cross-species transfer from non-human primates to humans using multi-dataset pretraining and full fine-tuning \citep{ryoo2025generalizable}. These works exploit shared structure across sessions, animals, or brain areas, but they do not directly test frozen-backbone transfer across heterogeneous calcium imaging datasets and species. For calcium imaging, transfer remains less systematically studied. POCO trains across calcium imaging datasets from zebrafish, mice, and \textit{C. elegans}, but mainly focuses on evaluation of forecasting performance within the datasets \citep{duan2025poco}. While POYO+ demonstrates transfer across mouse calcium datasets, it is mainly a supervised decoding framework and does not demonstrate cross-species transfer \citep{azabou2025multi}. Thus, whether a pretrained calcium backbone can transfer across tasks, datasets, and species remains underexplored.

\vspace{-0.3cm} 
\section{Method}
\vspace{-0.2cm} 

\begin{figure}[!t]
  \centering
  \includegraphics[width=\linewidth]{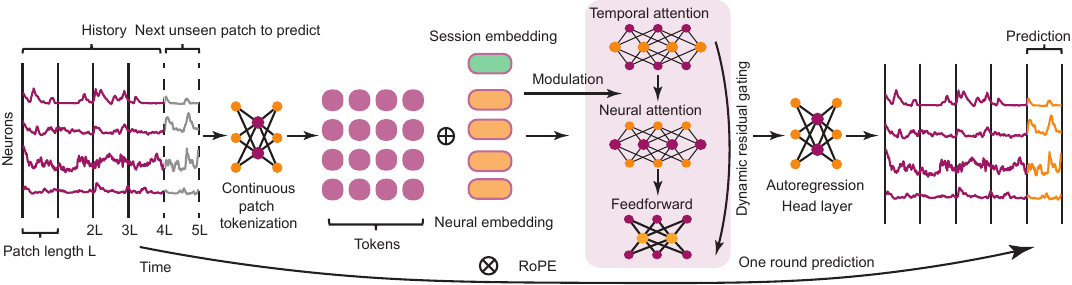}
  \caption{
  \textit{Pretraining pipeline for CAPT.}}
  \label{fig:fig1}
\end{figure}

Calcium imaging provides cellular-resolution measurements of neural population
dynamics, yet learning reusable representations remains challenging due to the
heterogeneity of neural recordings across datasets. We hypothesize that directly
modeling continuous calcium dynamics with autoregressive objectives can provide a
unified representation of neural population activity. Based on this intuition, CAPT models calcium dynamics in continuous space through a patch-based autoregressive Transformer. Each neuron trace is segmented into temporal patches, projected with neuron and session embeddings, and processed by population-level attention. By predicting future calcium activity directly, CAPT learns representations aligned with neural dynamics while enabling efficient adaptation to new datasets through transfer modules. (Figure \ref{fig:fig1}).

\subsection{Continuous patch tokenization}

Given a neural population segment, such as a trial, from session $s$, we denote it as
\begin{equation}
    \mathbf{X} \in \mathbb{R}^{N_s \times T},
\end{equation}
where $N_s$ is the number of recorded neurons and $T$ is the number of time
points. CAPT first divides each single-neuron trace into non-overlapping temporal
patches of length $L$. We highlight the effectiveness of temporal compression, i.e., $L>1$, through an ablation study in Appendix \ref{app:ablation}. The $p$-th patch of neuron $n$ is denoted as
\begin{equation}
    \mathbf{x}_{n,p}
    =
    \left[
    x_{n,(p-1)L+1},
    \ldots,
    x_{n,pL}
    \right]^\top
    \in \mathbb{R}^{L}.
\end{equation}

Subsequently, each continuous patch is directly projected into the model space:
\begin{equation}
    \mathbf{z}_{n,p}
    =
    \mathbf{W}_{\mathrm{in}}\mathbf{x}_{n,p}
    + \mathbf{e}_{n}
    + \mathbf{g}_{s}.
\end{equation}
Here, $\mathbf{W}_{\mathrm{in}}$ is the continuous patch projection that maps the
patch into the model dimension, $\mathbf{e}_{n}$ is the neuron embedding, and
$\mathbf{g}_{s}$ is the session embedding. In practice, temporal order is encoded
through rotary positional encoding (RoPE) in the temporal attention layers
\citep{su2024roformer}.

The projected continuous patches are processed by the Dual-axis Transformer (DAT)
backbone \citep{xu2026self}. The backbone factorizes population modeling into two
complementary directions: causal temporal attention models the history of each
neuron, while neuron-axis attention models population-level interactions at each
patch position. The backbone outputs patch-level neural representations
\begin{equation}
    \mathbf{h}_{n,p} \in \mathbb{R}^{D}.
\end{equation}

\subsection{Network architecture}

To further improve the adaptability of population representation learning, we introduce two lightweight architectural components: dynamic residual gating and neuron-aware QKV modulation.

\vspace{-0.1cm}
\paragraph{Dynamic residual gating.}
To adapt to heterogeneous neural dynamics, we replace fixed residual scaling with
dynamic residual gating. Specifically, for a
Transformer sub-layer $F(\cdot)$, the residual update is formulated as
\begin{equation}
    \mathbf{h}' =
    \mathbf{h}
    +
    g(\mathbf{h})F(\mathrm{LN}(\mathbf{h})),
\end{equation}
where $g(\mathbf{h})$ is a token-dependent scalar gate generated by a lightweight
two-layer MLP. Separate gates are applied to temporal attention, neuron-axis attention, and feed-forward branches. This adaptive residual mechanism allows different neural states to dynamically control the strength of representation updates while preserving the original residual structure.

\vspace{-0.1cm}
\paragraph{Neuron-aware QKV modulation.}
Neuron identity can provide useful priors for population interaction modeling.
We therefore modulate the query, key, and value projections in neuron-axis
attention using neuron embeddings:
\begin{equation}
    \mathbf{x}_n
    =
    W_x\mathbf{z}_n + U_x(\mathbf{e}_n),
    \qquad x\in\{q,k,v\},
\end{equation}
where $U_x$ denotes a low-rank adapter for the corresponding projection.
The modulation enables neuron-aware attention by allowing identity information
to influence both interaction weights and aggregated representations. (Appendix~\ref{app:ablation})

\subsection{Autoregressive pretraining in continuous space}

CAPT is pretrained to predict future continuous calcium activity from past
activity. Given the patch-level representation $\mathbf{h}_{n,p}$, the prediction
head outputs the future residual patch:
\begin{equation}
    \widehat{\mathbf{r}}_{n,p+1}
    =
    \mathbf{W}_{\mathrm{out}}\mathbf{h}_{n,p}
    \in \mathbb{R}^{L}.
\end{equation}
To improve autoregressive forecasting stability, CAPT adopts residual-last
prediction, where future activity is modeled as the change relative to the last
observed value. Specifically, given the latest observed activity
$x_{n,t}$, the residual target is defined as
\begin{equation}
    r_{n,t+k}=x_{n,t+k}-x_{n,t},
    \qquad k=1,\ldots,L .
\end{equation}
During training and inference, the predicted residual is converted back to the calcium space by:
\begin{equation}
    \widehat{x}_{n,t+k}
    =
    x_{n,t}+\widehat{r}_{n,t+k}.
\end{equation}

The autoregressive objective is defined as the mean squared error over future
residual patches:
\begin{equation}
    \mathcal{L}_{\mathrm{AR}}
    =
    \frac{1}{|\Omega|}
    \sum_{(n,p,\ell)\in\Omega}
    \left(
    \widehat{r}_{n,p+1,\ell}
    -
    r_{n,p+1,\ell}
    \right)^2 .
\end{equation}
During training, future residual patches are predicted in parallel under teacher
forcing, where previous patches are used to predict subsequent patches.

\vspace{-0.1cm}
\subsection{Behavior decoding head}
\vspace{-0.1cm}

In all decoding experiments, we freeze the pretrained CAPT backbone and train a
task-specific decoder on top of the neural representations. Given
the frozen backbone output $H$, the decoder predicts a behavior trace
$\widehat{\mathbf{Y}}\in\mathbb{R}^{C_b\times T}$, where $C_b$ is the number of
behavior dimensions and $T=PL$.

We use a nonlinear low-rank per-neuron readout to aggregate information across
neurons. For each neuron representation, a gated feature is computed as
\begin{equation}
    q_{n,p,o,r}
    =
    \left(\mathbf{a}^{u}_{o,r}\right)^\top \mathbf{h}_{n,p}
    \cdot
    \mathrm{GELU}
    \left(
    \left(\mathbf{a}^{v}_{o,r}\right)^\top \mathbf{h}_{n,p}
    \right),
\end{equation}
where $o=1,\ldots,C_bL$ indexes the within-patch behavior outputs and
$r=1,\ldots,R_{\mathrm{dec}}$ is the low-rank index. The patch-level output is then obtained by
a neuron-wise weighted aggregation:
\begin{equation}
    \widehat{y}_{p,o}
    =
    \sum_{n=1}^{N_s}
    \sum_{r=1}^{R_{\mathrm{dec}}}
    q_{n,p,o,r} B_{o,\mathrm{id}(n),r}
    + b_o .
\end{equation}
Here, $\mathrm{id}(n)$ denotes the global neuron index and
$B_{o,\mathrm{id}(n),r}$ and $b_o$ are learnable projection weights and bias. The
$C_bL$ outputs are reshaped into $C_b$ behavior channels at the original temporal
resolution, followed by an optional depthwise temporal convolution for smoothing.

The decoder is trained with MSE:
\begin{equation}
    \mathcal{L}_{\mathrm{dec}}
    =
    \frac{1}{|\Omega_y|}
    \sum_{(c,t)\in\Omega_y}
    \left(
    \widehat{y}_{c,t}
    -
    y_{c,t}
    \right)^2,
\end{equation}
where $|\Omega_y|$ is the number of entries used to compute the behavior loss.

\vspace{-0.1cm}
\subsection{Transfer setup}
\label{sec:transfer}
\vspace{-0.1cm}

CAPT is first pretrained on the source dataset using the
self-supervised autoregressive objective:
\begin{equation}
    \theta^\star
    =
    \arg\min_{\theta}
    \mathbb{E}_{\mathbf{X}\sim\mathcal{D}_{\mathrm{src}}}
    \left[
    \mathcal{L}_{\mathrm{AR}}(\mathbf{X};\theta)
    \right].
\end{equation}

For a target dataset, transfer is performed in two separate ways.

\paragraph{Forecasting adaptation by tuning new embeddings.}

We expand the neuron
and session embedding tables with target-specific entries for new sessions and neurons. The patch
projection, backbone, and autoregressive prediction head are frozen.
Only the new target neuron and session embeddings are optimized using the same
autoregressive forecasting objective:
\begin{equation}
    \psi_{\mathrm{tgt}}^\star
    =
    \arg\min_{\psi_{\mathrm{tgt}}}
    \mathbb{E}_{\mathbf{X}\sim\mathcal{D}_{\mathrm{tgt}}}
    \left[
    \mathcal{L}_{\mathrm{AR}}
    \left(
    \mathbf{X};
    \theta^\star,
    \psi_{\mathrm{tgt}}
    \right)
    \right].
\end{equation}
This step evaluates whether the frozen CAPT backbone can support neural
population forecasting on a new dataset by learning only
dataset-specific embeddings. Under this transfer protocol, increasing the number of target sessions does not provide additional shared trainable capacity, and fine-tuning performance reflects whether the frozen backbone can be reused through per-session adaptation.

\paragraph{Behavior decoding by tuning only the decoder.}
After the target neuron and session embeddings are adapted, we train only the behavior
decoder:
\begin{equation}
    \phi_{\mathrm{dec}}^\star
    =
    \arg\min_{\phi_{\mathrm{dec}}}
    \mathbb{E}_{(\mathbf{X},\mathbf{Y})\sim\mathcal{D}_{\mathrm{tgt}}}
    \left[
    \mathcal{L}_{\mathrm{dec}}
    \left(
    \mathbf{Y},
    g_{\phi_{\mathrm{dec}}}
    \left(
    f_{\theta^\star,\psi_{\mathrm{tgt}}^\star}
    \left(
    \mathbf{X}
    \right)
    \right)
    \right)
    \right].
\end{equation}
Here, $f_{\theta^\star,\psi_{\mathrm{tgt}}^\star}$ denotes the frozen
embedding-adapted CAPT backbone, and $g_{\phi_{\mathrm{dec}}}$ denotes the
trainable per-neuron behavior decoder. 

This two-step protocol separates neural
population adaptation from task adaptation: forecasting transfer only updates new
embeddings, while behavior decoding only updates the decoder. Therefore, the
experiments directly test whether a single continuous autoregressive backbone
can provide reusable representations across datasets, experimental paradigms,
and species.

\section{Benchmark and Evaluation}

\subsection{Datasets}

For pretraining, we use a large-scale mouse calcium imaging dataset
from Tseng et al., used to evaluate held-in and held-out performance within the pretraining distribution following CalM.

To evaluate cross-dataset and cross-species transfer, we further use eight
independent multi-animal, multi-session calcium imaging datasets collected from different laboratories and
model organisms, including three mouse datasets, three larval zebrafish datasets,
and two \textit{C. elegans} datasets. These datasets differ in species, imaging
conditions, neuron counts, sampling rates, behavioral variables, and experimental
paradigms, providing a broad benchmark for testing whether a single frozen CAPT
backbone can be reused beyond the original pretraining distribution (Appendix \ref{app:dataset}).

\subsection{Tasks for performance measurement}

As mentioned, we evaluate CAPT on two predictive tasks.

\begin{figure}[!t]
  \centering
  \includegraphics[width=\linewidth]{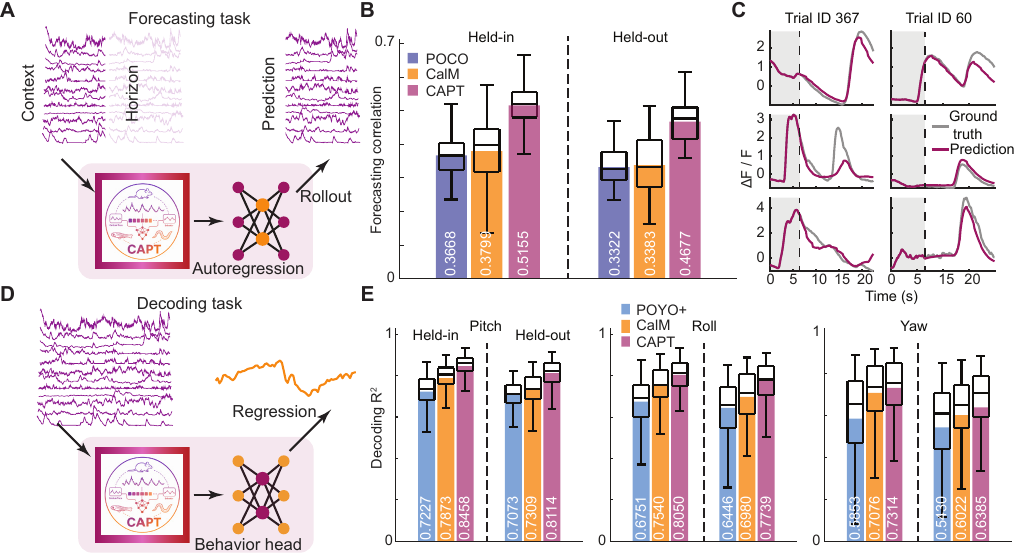}
  \caption{
  \textit{Performance of CAPT on pretraining dataset.} (A) Illustration of CAPT on forecasting task. (B) Neural population forecasting performances on the pretraining mouse dataset. (C) Example trace visualization. (D) Illustration of CAPT on behavior decoding tasks. (E) Behavior decoding performances on the pretraining mouse dataset.
  }
  \label{fig:fig2}
\end{figure}

\noindent\textbf{Neural population forecasting.}
Given a context window of calcium population activity, CAPT autoregressively predicts the horizon of future continuous traces (Figure \ref{fig:fig2}A). We report Pearson correlation between predicted and ground-truth traces.

\noindent\textbf{Behavior decoding.}
Given neural activity from a trial or pseudo-trial, CAPT decodes continuous behavioral variables (Figure \ref{fig:fig2}D). We report $R^2$ for each behavior variable.

\begin{table}[t]
\centering
\caption{Summary of datasets used for pretraining and transfer evaluation.}
\label{tab:dataset_summary}
\resizebox{\linewidth}{!}{
\begin{tabular}{lllllll}
\toprule
Dataset & Species & \# Subjects & \# Sessions & \# Neurons & $f_s$ & $\mathrm{Ca^{2+}}$ indicator \\
\midrule
Tseng et al. \citep{tseng2022shared} & Mouse & 8 & 286
& 273,770 
& 6 Hz & GCaMP6s \\

Sun et al. \citep{sun2025learning} & Mouse & 13 & 52
& 212,464 
& 10 Hz & GCaMP6f \\

CaLiAli \citep{vergara2025comprehensive} & Mouse & 3 & 3
& 988 
& 15.6 Hz & GCaMP6f \\

Park et al. \citep{park2024decoding} & Mouse & 3 & 3
& 9,619  
& 7.8 Hz & GCaMP6s\\

Lavian et al. \citep{lavian2025visual} & Larval zebrafish & 21 & 45
& 140,311 
& 2-5 Hz & GCaMP6s \\

Brysch et al. \citep{brysch2019functional} & Larval zebrafish & 3 & 29
& 4,026 
& 2 Hz & GCaMP6f \\

Palieri et al. \citep{palieri2024preoptic} & Larval zebrafish & 5 & 5
& 20,480 
& 2 Hz & GCaMP6s \\

Atanas et al. \citep{atanas2023brain} & \textit{C. elegans} & 80 & 80
& 10,782 
& 1.67 Hz & GCaMP7f\\

Copper boundary \citep{baskoylu2026copper} & \textit{C. elegans} & 27 & 27
& 3,984 
& 1.67 Hz & GCaMP7f\\
\bottomrule
\end{tabular}
}
\end{table}

\subsection{Baselines}

For task-specialized methods, we use POCO for neural population forecasting and POYO+ for behavior decoding. In addition, we compare with CalM for a general-purpose calcium modeling. The details of the baseline and CAPT setup can be found in the Appendix \ref{app:baseline} and \ref{app:model}.

\vspace{-0.1cm} 

\section{Experiments and Results}

\vspace{-0.1cm}
For all datasets, we split trials or pseudo-trials into training, validation,
and testing sets with a ratio of 70\%, 15\%, and 15\%, respectively whenever
applicable. The validation set is used for model selection, and all reported results are computed on the test set. All the transfer experiments, as well as held-out part from pretraining data, follow transfer setup, where the pretrained CAPT backbone is frozen and adaptation modules, i.e., session and neuron embedding and decoding head, are updated.

\vspace{-0.1cm}
\subsection{Performance on the pretraining dataset}
\vspace{-0.1cm}

We first evaluate whether CAPT can improve the source-domain performance. CAPT outperforms both POCO and CalM on held-in and held-out
sessions (Figure \ref{fig:fig2} B, C), indicating that continuous autoregressive modeling provides a strong
alternative to discrete-token autoregression or POYO-augmented MLP method for calcium trace forecasting.

For behavior decoding, CAPT achieves stronger decoding performance than the
baselines across behavioral variables, demonstrating the effectiveness of learned representations (Figure \ref{fig:fig2} E).

\begin{figure}[!t]
  \centering
  \includegraphics[width=\linewidth]{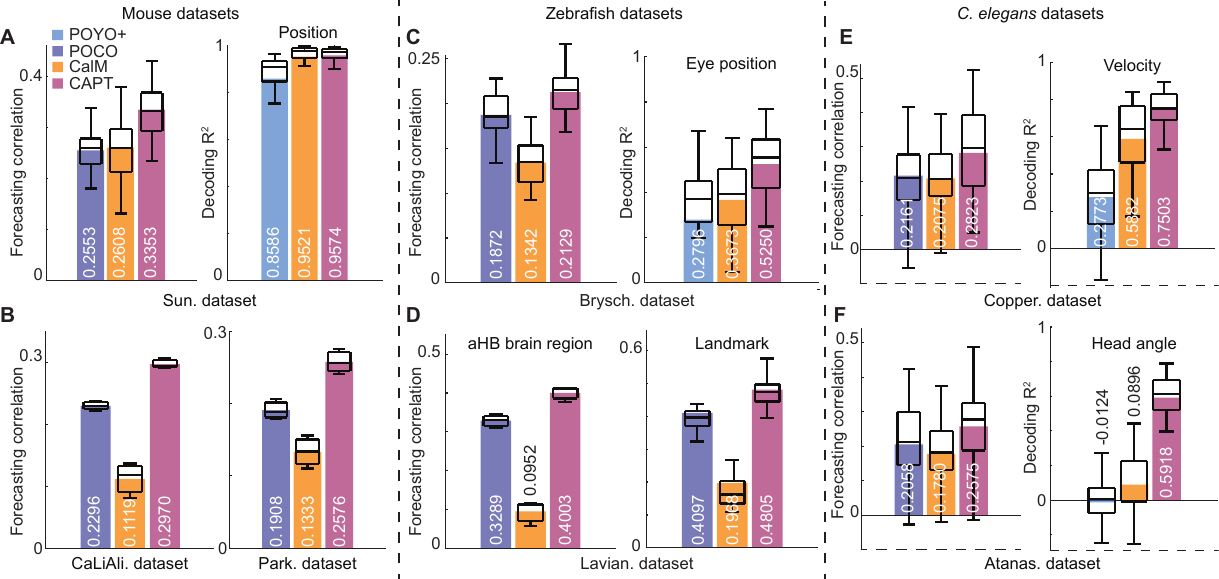}
  \caption{
  \textit{Performance of CAPT on diverse transfer datasets.} (A) (B) Neural population forecasting and behavior decoding performances on Sun., CaLiAli. and Park. mouse datasets. (C) (D) Neural population forecasting and behavior decoding performance on Brysch. and Lavian. zebrafish datasets. (E) (F) Forecasting and decoding performance on Copper. and Atanas. \textit{C. elegans} datasets.
  }
  \label{fig:fig3}
\end{figure}

\vspace{-0.1cm}
\subsection{Performance on cross-lab mouse datasets}
\vspace{-0.1cm}

We next evaluate whether CAPT transfers to independent mouse datasets collected
outside the pretraining distribution. This setting tests cross-dataset
generalization under changes in experimental paradigm, neural population, and
behavioral variables.

As shown in Figure \ref{fig:fig3}A, B, CAPT consistently outperforms the applicable
forecasting baselines on the cross-lab mouse datasets. Since the backbone is
frozen and only new neuron and session embeddings are updated, these results show
that CAPT can reuse the pretrained continuous backbone for new mouse calcium
recordings and is more stable than CalM.

\subsection{Performance on cross-species datasets}

We further test CAPT in a more challenging cross-species transfer setting,
including larval zebrafish and \textit{C. elegans} datasets. Compared with
cross-lab mouse transfer, these datasets introduce larger shifts in organism,
recording scale, sampling rate, and experimental paradigm.

On larval zebrafish datasets, CAPT achieves better forecasting performance than
the baselines across different recording settings (Figure \ref{fig:fig3}C). In settings where pretrained POCO is unavailable, CAPT remains directly applicable and achieves strong performance (Appendix \ref{app:result}).

On \textit{C. elegans} datasets, CAPT also outperforms applicable baselines
for neural forecasting and achieves superior performance
on available behavioral variables, such as head-angle-related
signals and velocity, where POYO+ and CalM show degradation (Figure \ref{fig:fig3}E, F), which is not simply caused by hyperparameter selection as in-domain pretrained POYO+ performs much better (Appendix \ref{app:result}).

These results suggest that a mouse-pretrained continuous autoregressive
backbone can surprisingly provide useful representations for compact nervous systems
with very different anatomical and functional organizations.

\vspace{-0.3cm}
\subsection{CAPT provides biologically meaningful embeddings}
\vspace{-0.1cm}

More importantly, we examine whether CAPT embeddings capture
biologically meaningful structure in transfer datasets, based
on NeuroPAL annotations of \textit{C. elegans} datasets, which provide cell-identity labels for a subset of recorded neurons. We select POCO as the baseline due to the second-best performance of this self-supervised method.

\begin{figure}[!t]
  \centering
  \includegraphics[width=\linewidth]{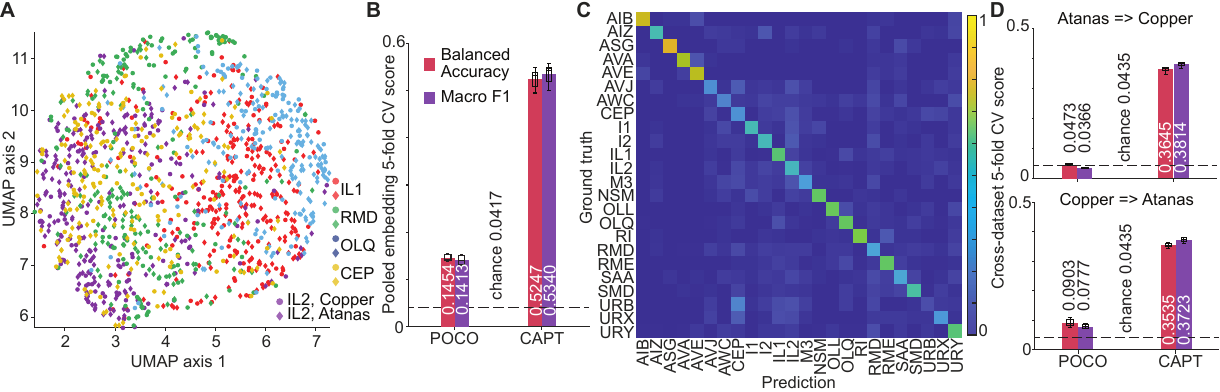}
  \caption{
  \textit{Neural embedding analysis with NeuroPAL information.} (A) UMAP Visualization of top-5  most abundant NeuroPAL-labelled embeddings. (B) Classification performances on pooled embedding of Copper. and Atanas datasets. (C) Confusion matrix of CAPT. (D) Cross-dataset classification performances.
  }
  \label{fig:fig4}
\end{figure}

We first visualize the learned neuron embeddings with UMAP \citep{mcinnes2018umap}.
As shown in Figure \ref{fig:fig4}A, neurons with different NeuroPAL identities form
structured distributions in the CAPT embedding space, suggesting that the model
captures identity-related organization rather than dataset-specific
variation, while the visualization of POCO embeddings appears to be more randomly distributed with the same UMAP parameter (Appendix \ref{app:result}).

To further quantify this structure, we perform cell-identity classification using pooled neuron embeddings with abundant NeuroPAL labels (number > 200) with a linear SVM. CAPT clearly outperforms POCO (Figure \ref{fig:fig4}B, C, Appendix \ref{app:result}), showing that CAPT embeddings preserve stronger cell-identity information. We further test a more representative and challenging cross-dataset setting, where the classifier is trained on one \textit{C. elegans} dataset and evaluated on the other. CAPT again achieves substantially stronger performance, indicating that its embedding space is shared across datasets rather than being restricted to a single recording, while the performance of POCO appears to be chance level (Figure~\ref{fig:fig4}D). The shared space arises from adaptation into the frozen pretrained backbone as retraining CAPT independently on each dataset drops cross-dataset classification to chance (Table~\ref{tab:neuropal_retrain}). We also demonstrate that this link cannot be discovered with solely trace statistics (Appendix \ref{app:result}).

Together, these results show that CAPT not only improves forecasting and decoding
performance, but also learns a shared functional embedding space that can be linked to
biological cell identity within different transfer \textit{C. elegans} datasets with even different data splitting setup (Appendix \ref{app:dataset}).

\vspace{-0.3cm}
\section{Conclusion}
\vspace{-0.1cm}

In this work, we introduce CAPT, a continuous autoregressive transformer able to transfer to new datasets by updating only adaptation modules after pretraining on a large-scale mouse calcium imaging dataset. \textbf{Across neural population forecasting and behavior decoding tasks, CAPT consistently outperforms specialized and general-purpose baselines on mouse, larval zebrafish, and \textit{C. elegans} datasets with distinct experimental paradigms.} Beyond predictive performance, in \textit{C. elegans} datasets where NeuroPAL annotations provide cell identity information, \textbf{CAPT provides a shared functional embedding space across datasets and captures anatomical, identity-related structure.} These results suggest that CAPT provides an effective framework for reusing a single pretrained model across different neural populations, laboratories, experimental paradigms, and species, and may serve as a step toward future neural foundation models for calcium imaging datasets.

Several limitations remain. (1) CAPT is pretrained on a single mouse calcium
imaging dataset. Multi-dataset pretraining across paradigms and species should
improve the robustness and generality of the learned representations. (2) The
performance gain under transfer is generally smaller than that on the
pretraining dataset, suggesting that the backbone remains partially constrained
by the source distribution. More diverse pretraining data may help close this
gap. (3) The tested methods generally perform better on trial-aligned datasets
than on recordings without clear trial structure, indicating that pseudo-trial
segmentation may be insufficient for spontaneous activity. Stimulus or
environment recordings may be needed in such settings. (4) The mechanism
underlying cross-species transfer remains only partially understood. Further
analyses of attention structure, embedding geometry, and biological modalities
are needed to clarify how CAPT generalizes across species.

A broader goal of such transferable models is to enable scientific discovery beyond predictive performance. In this context, NeuroPAL-identified neurons can be linked to connectomic and transcriptomic resources, providing a unique opportunity to relate functional neural activity to circuit and molecular context. Our NeuroPAL analysis suggests that CAPT embeddings are not only useful for forecasting and decoding, but may also provide a functional coordinate system that aligns neural activity with cell identity across datasets. Together, these observations suggest that future CAPT-style models may provide a bridge from neural computation to biological structures across species, which remain fundamental questions in the neuroscience community.

\section*{AI use statement}

In this work, we used generative AI tools to assist with literature search and summarization, code development and debugging, interpretation and organization of experimental results, and language and LaTeX editing of the manuscript. All AI-assisted suggestions, analyses, and code were critically reviewed by the authors; relevant scientific claims were checked against the original literature, and AI-assisted code was verified through inspection and empirical testing. We take full responsibility for the final content of this work, including all text, claims, analyses, and artifacts produced with the aid of generative AI.

\section*{Ethics Statement}
This work uses only previously published, publicly available animal calcium
imaging datasets; no new animal experiments were performed and no human
subjects or human neural data are involved. All datasets are used in accordance
with their original licenses and access terms.

This work proposes a continuous autoregressive model for calcium-imaging
population activity to learn transferable neural representations across
datasets and species, and to improve downstream analyses such as forecasting
and behavior decoding. It may reduce task-specific modeling and support more
unified and data-efficient neuroscience research. Potential risks include
over-interpretation or misuse of learned representations, and privacy concerns
if similar approaches are applied to human neural data. This paper makes no
clinical claims; any extension to human recordings should require ethical
review, informed consent, strong de-identification and access control, and
careful reporting of limitations and uncertainty.

\section*{Reproducibility Statement}
All datasets used in this work are publicly available and are described in
Appendix~\ref{app:dataset}, including per-dataset preprocessing, causal EMA
smoothing, normalization fitted on the training split only, and the exact
trial / pseudo-trial construction and 70/15/15 split procedure for each of the
nine datasets. Architecture and hyperparameters for CAPT are given in
Appendix~\ref{app:model} (Tables~\ref{tab:capt_backbone} and~\ref{tab:capt_decoder}),
and baseline configurations and fine-tuning protocols for POCO, POYO+, and CalM
in Appendix~\ref{app:baseline} (Tables~\ref{tab:poco}--\ref{tab:calm_decoder}).
The transfer protocol is specified in Section 3.5. Evaluation metrics and the
hierarchical neuron/trial/session aggregation used for all reported numbers are
defined in Appendix~\ref{app:dataset}; statistical comparisons and exact
$p$-values are in Table~\ref{tab:statistical_test}. Compute is reported in
Appendix F. Anonymized code for preprocessing, pretraining, transfer, and the
NeuroPAL analyses is available at
\url{https://anonymous.4open.science/r/project_code_2026-0AC8/}.

\bibliographystyle{iclr2027_conference}
\bibliography{reference}


\newpage
\appendix

\section{Dataset}
\label{app:dataset}
In this work, we use a collection of calcium imaging datasets spanning three species (mouse, larval zebrafish, and \textit{C. elegans}) and covering diverse experimental paradigms. In total, the datasets comprise 676,424 neurons and 530 sessions. All neural traces are smoothed and z-score normalized, while behavioral variables are z-score normalized before training. To prevent leakage across train/validation/test splits and temporal leakage within trials or pseudo-trials, neural smoothing is performed with a causal exponential moving average (EMA) with $\alpha=0.15$, and the EMA state is not propagated across split boundaries. For normalization, the mean and standard deviation are estimated only from the training split and the fitted parameters are directly applied to the corresponding validation and test splits. All reported metrics are aggregated hierarchically to avoid domination by large sessions. For neural population forecasting, Pearson correlation is computed per neuron within each trial over prediction time points, averaged across neurons to obtain a trial-level score, then averaged across trials within each session and finally across sessions. For behavior decoding, $R^2$ is computed separately for each behavior channel within each trial, followed by the same trial-level and session-level aggregation.

\textbf{Mouse}

\textbf{Tseng. dataset}: The dataset used for pretrain from \citep{tseng2022shared} comprises calcium recordings of 286 sessions from 8 mice, totaling 273,770 neurons recorded from 6 brain regions and 2 cortical layers during a decision-making task. In the task, mice are placed in a Y-maze, where they receive a visual cue (black or white) in the stem section and subsequently make a left or right turn in the arm section according to the cue. The cue-choice mapping rule is switched every over 100 trials. The original multi-plane imaging data are acquired at 30 Hz. In our analysis, neurons from different planes are treated as being sampled simultaneously, resulting in an effective sampling rate of 6 Hz. The three-dimensional velocity from the spherical treadmill (pitch, roll, and yaw) is used as the behavioral variable for the decoding task. The dataset is partitioned into held-in and held-out subset by mouse subjects. The held-in dataset is used for pretrain (6 subjects, 189 sessions, 197,704 neurons) and held-out for within-dataset finetune (2 subjects, 97 sessions, 76,066 neurons). Within each session, trials are split into training, validation, and test sets with a ratio of 70:15:15. This splitting ratio is consistently applied across all datasets used in this paper. 

\textbf{Sun. dataset}: The mouse dataset from \citep{sun2025learning} consists of calcium imaging recordings of the hippocampus with a sampling rate of 10 Hz. It is collected from multiple mice across days during a navigation task. In this task, mice navigate a linear track and encounter sequences of sensory cues. Depending on the cue sequence, mice are required to infer the trial type and produce licking behavior at specific positions. To facilitate model training, trials shorter than 64 frames are excluded, while those longer than 200 frames are truncated to 200 frames. Neurons from different sessions of the same subject are treated as distinct. The final dataset comprised 52 sessions from 13 subjects, totaling 212,464 neurons and 7,564 trials. Each trial contains an average of approximately 133 frames. The position of the mouse was used as behavior for decoding task.

\textbf{CaLiAli dataset}: The mouse dataset CaLiAli from \citep{vergara2025comprehensive} consists of one-photon calcium imaging recordings of hippocampal CA1 place cells from freely moving mice on a linear track. The dataset contains 3 sessions from 3 subjects and 988 neurons in total. For each subject, the four daily recordings are first separated according to the frame counts provided in the dataset, ensuring that pseudo-trials did not cross day boundaries. Within each day, the continuous recording is divided into non-overlapping temporal blocks of 1600 neural frames. In each block, continuous temporal splits which contain the first 70\% of frames are assigned to training, the following 15\% to validation, and the final 15\% to testing. Within each split, pseudo-trials are generated using sliding windows of 80 frames with a stride of 10 frames, ensuring that no pseudo-trial crossed split or day boundaries. This procedure yields 11176 pseudo-trials in total, with each pseudo-trial containing 80 frames.

\textbf{Park. dataset}: The mouse dataset from \citep{park2024decoding} consists of two-photon calcium imaging recordings collected from head-fixed mice freely running on an air-lifted spherical treadmill. Neural activity was recorded from somatosensory cortex at a sampling rate of 7.8 Hz. This dataset comprises 3 sessions from 3 subjects, totaling 9,619 neurons. Pseudo-trials are constructed from its continuous recordings. Each session is divided into non-overlapping temporal blocks of 800 neural frames, and each block is randomly assigned to training, validation, and test splits. Within each split, pseudo-trials are generated using sliding windows of 80 frames with a stride of 20 frames. This procedure yields 1,221 pseudo-trials in total, with each pseudo-trial containing 80 frames.

\textbf{Larval zebrafish}

\textbf{Lavian. dataset}: The larval zebrafish dataset \citep{lavian2025visual} contains three parts: whole brain, aHB and landmark.  

The whole-brain subset consists of light-sheet calcium imaging recordings from head-fixed larval zebrafish during a whole-field visual-motion stimulation paradigm. Visual stimuli consisted of translational motion patterns presented along multiple directions, and neural activity is recorded across the whole brain at 2 Hz. This subset contains 15 sessions from 12 subjects, comprising 493,695 raw neurons in total. To facilitate model training, only representative neurons are chosen by K-means clustering to reduce each session to 4,096 neurons, resulting in 61,440 neurons in total. Stimulus-aligned trials with a 16-frame context followed by a 16-frame horizon are reconstructed. This procedure produced 1,200 stimulus-aligned trials.

The aHB subset is collected under the same visual-motion stimulation paradigm, with imaging restricted to the anterior hindbrain at 5 Hz. This subset contains 3 sessions from 3 subjects, comprising 4,839 neurons in total. Stimulus-aligned trials with a 40-frame context followed by a 24-frame horizon are reconstructed. This procedure produced 204 stimulus-aligned trials.

The landmark subset consists of two-photon calcium imaging recordings at 3 Hz from head-fixed larval zebrafish during a landmark-position stimulation task. In this task, a visual landmark was presented at different azimuthal positions in the frontal visual field. This subset contains 27 sessions from 6 subjects, comprising 74,032 neurons in total. Landmark-aligned trials with a 40-frame context followed by a 24-frame horizon are constructed according to landmark onset. This procedure produced 1,920 landmark-aligned trials.

\textbf{Brysch. dataset}: The larval zebrafish dataset from \citep{brysch2019functional} consists of calcium imaging recordings collected during an optokinetic stimulation task, in which visual stimuli are used to evoke controlled eye movements. The visual stimulus is discretized into position bins, and within each bin, velocity is systematically varied. We retain recording sessions with more than 100 neurons. Each session is divided into non-overlapping temporal blocks of 1000 neural frames, and each block is randomly assigned to training, validation, and test splits. Within each split, pseudo-trials are generated using sliding windows of 64 frames with a stride of 16 frames, resulting in 29 sessions from 3 subjects, with 4,026 neurons and 4,635 pseudo-trials in total. The binocularly averaged eye position is used as the behavioral variable.

\textbf{Palieri. dataset}: The larval zebrafish dataset from \citep{palieri2024preoptic} consists of light-sheet whole-brain calcium imaging recordings from head-restrained larval zebrafish during a homeostatic stimulation paradigm. We use the short-protocol temperature task in which repeated temperature stimulation blocks are delivered. This subset contains 5 sessions from 5 subjects, comprising 313,319 raw neurons in total. Neurons are downsampled using the same clustering-based strategy, resulting in 20,480 in total and 4,096 neurons per session. Each continuous session was divided into three non-overlapping blocks of 1600 frames. Within each block, temporal continuous splits are assigned to training, validation and testing. Within each split, pseudo-trials are generated using sliding windows of 80 frames with a stride of 20 frames. This procedure resulted in 1,065 pseudo-trials in total. 

\textbf{\textit{C. elegans}}

\textbf{Atanas. dataset}: The \textit{C. elegans} dataset from \citep{atanas2023brain} consists of calcium imaging recordings from freely moving \textit{C. elegans} during spontaneous behavior. Neural activity is recorded at an effective sampling rate of approximately 1.67 Hz. The dataset contains 80 sessions, totaling 10,782 neurons. Pseudo-trials are constructed by first dividing each session into non-overlapping blocks of 800 frames. Within each block, continuous temporal splits are assigned to training, validation, and test splits. Within each split, pseudo-trials were generated using sliding windows of 80 frames with a stride of 10 frames. This procedure yielded 9,145 pseudo-trials in total. The head angle is used as the behavioral variable for decoding task.

\textbf{Copper boundary dataset}: The \textit{C. elegans} Copper dataset from \citep{baskoylu2026copper} consists of calcium imaging recordings from freely moving \textit{C. elegans} during an aversive copper-boundary encounter paradigm. \textit{C. elegans} expressing the NeuroPAL transgene are recorded while exploring circular arenas surrounded by copper-containing agar. Each recording contains 1,600 confocal calcium-imaging frames, corresponding to a sampling rate of 1.67 Hz. The dataset contains 27 sessions from 27 animals, totaling 3,984 neurons. For each session, the continuous recording is divided into 20 non-overlapping pseudo-trials of 80 frames, resulting in 540 pseudo-trials in total. This setup is used to evaluate the transfer ability of CAPT under limited-trial settings. The velocity is used as the behavioral variable for decoding task.

Both \textit{C. elegans} datasets contain NeuroPAL-based cell-identity annotations for a subset of recorded neurons \citep{yemini2021neuropal}. In our processed data, NeuroPAL identities are available for 2,159 of 10,782 neurons in the Atanas dataset and for all 3,984 neurons in the Copper dataset, resulting in 6,143 NeuroPAL-labeled neurons across the two datasets. These annotations are used only for downstream multimodal biological analyses.

\section{Baseline and training details}
\label{app:baseline}

\textbf{POCO}

For POCO \citep{duan2025poco} on the multi-session forecasting task, the model requires trials to have a fixed context and forecasting horizon. Accordingly, all trials are truncated to 64 time steps (40 for context and 24 for forecasting), and trials shorter than this length are discarded. This fixed-length configuration is kept consistent between pretraining and fine-tuning. We adopt the sliding window hyperparameter in POCO with a stride of 64 to avoid splitting or concatenating trials during training and evaluation. Hyperparameters are selected via a broad search based on the validation Pearson correlation coefficient. For each experiment, models are trained for 200 epochs, and the checkpoint with the best validation correlation is selected. Detailed hyperparameter settings for pretraining are provided in Table~\ref{tab:poco}. During fine-tuning, we freeze the POYO backbone and the conditioning MLP, and only update the neural and session embeddings.

\begin{table}[!htbp]
  \caption{Hyperparameters used for training POCO model}
  \label{tab:poco}
  \centering
  \begin{tabular}{ll}
    \toprule
    Hyperparameter & Value \\
    \midrule
    MLP hidden size & 1024 \\
    Decoder layers & 1 \\
    Decoder hidden size & 128 \\
    Number of heads & 16 \\
    Latent tokens & 8 \\
    FFN hidden size & 1024 \\
    Compression Factor & 8 \\
    \bottomrule
  \end{tabular}
\end{table}

\textbf{POYO+}

For POYO+ \citep{azabou2025multi} on the multi-session decoding task, we conduct a broad hyperparameter search on a small subset of 9 sessions from the pretraining dataset. The best configuration selected based on validation $R^2$ is then applied to the full pretrain dataset. Only the learning rate is adjusted to ensure stable and effective training. Detailed hyperparameters are provided in Table~\ref{tab:poyo}. For cross-dataset fine-tuning, we only adjust the sequence length and latent step based on real time to match sampling rates across different datasets. The pretrain backbone is kept frozen across all fine-tuning experiments. For within-dataset fine-tuning, the decoder head is frozen, and only the neural and session embeddings are updated. For cross-dataset fine-tuning, a new decoder head is initialized to match the corresponding behavioral variables, and both the decoder head and embeddings are jointly updated. 

\begin{table}[!htbp]
  \caption{Hyperparameters used for training POYO+ model}
  \label{tab:poyo}
  \centering
  \begin{tabular}{ll}
    \toprule
    Hyperparameter & Value \\
    \midrule
    Embedding Dimension & 128 \\
    Head Dimension & 64 \\
    Number of Latents & 32 \\
    Sequence length & 2.0 \\
    Latent Step & 0.125 \\
    Depth & 24 \\
    Self-attention Heads & 8 \\
    Cross-attention Heads & 4 \\
    FFN Dropout & 0.2 \\
    Linear Dropout & 0.4 \\
    Attention Dropout & 0.2 \\
    \bottomrule
  \end{tabular}
\end{table}

\textbf{CalM}

For CalM \citep{xu2026self} on multi-session forecasting and decoding tasks, we first perform hyperparameter selection on a 12-session dataset, choosing the configuration that achieves the best validation correlation for the forecasting task. The NQ module is pretrained on the pretraining dataset and subsequently used to tokenize other fine-tuning datasets. The DAT module is pretrained on the same pretraining dataset. For forecasting fine-tuning, we freeze the pretrained DAT backbone and update only neuron and session embeddings. For decoding tasks that introduce new behavioral variables, we initialize and train only a new decoding head while keeping the backbone fixed. For within-dataset held-out decoding fine-tuning, we update only the new neuron-specific rows in the decoding head corresponding to the held-out neurons. We train for 200 epochs during pretraining and within-dataset fine-tuning of DAT, and for all decoding tasks. For cross-dataset fine-tuning of DAT, we extend training to 400 epochs to ensure convergence. Detailed hyperparameters are provided in Table~\ref{tab:calm_tseng}, ~\ref{tab:axialar_actual} and \ref{tab:calm_decoder}. 

\begin{table}[!htbp]
  \caption{Hyperparameters used for CalM (NQ) model}
  \label{tab:calm_tseng}
  \centering
  \begin{tabular}{ll}
    \toprule
    Hyperparameter & Value \\
    \midrule
    Embedding Dimension & 512 \\
    Codebook Size & 128 \\
    Encoder / Decoder Layers & 4 \\
    Attention Heads & 4 \\
    Discretization Window / Overlap & 4 \\
    EMA Decay & 0.99 \\
    Max AR Horizon & 4 \\
    \bottomrule
  \end{tabular}
\end{table}

\begin{table}[!htbp]
  \caption{Hyperparameters used for CalM (DAT) model}
  \label{tab:axialar_actual}
  \centering
  \begin{tabular}{ll}
    \toprule
    Hyperparameter & Value \\
    \midrule
    Model Dimension & 512 \\
    Layers & 6 \\
    Attention Heads & 8 \\
    FFN Dimension & 2048 \\
    Scheduled sampling probability & 0.6 \\
    Neighborhood replacement probability & 0.1 \\
    \bottomrule
  \end{tabular}
\end{table}

\begin{table}[!htbp]
  \caption{Hyperparameters used for CalM (downstream decoder head) model}
  \label{tab:calm_decoder}
  \centering
  \begin{tabular}{ll}
    \toprule
    Hyperparameter & Value \\
    \midrule
    Rank & 16 \\
    Head dropout & 0.6 \\
    Temporal convolution kernel & 9 \\
    \bottomrule
  \end{tabular}
\end{table}

\section{Model and hyperparameters}
\label{app:model}

For CAPT, we pretrain a continuous patch-based autoregressive backbone on the multi-session forecasting task. Continuous calcium traces are divided into non-overlapping temporal patches, and the model predicts the next continuous patch directly in the original trace space. We use mean squared error as the forecasting objective. During pretraining, we use teacher-forced next-patch prediction. During evaluation, CAPT is evaluated by true autoregressive rollout from a fixed context window.

Following the same protocol as CalM, we perform hyperparameter selection on a 12-session subset and choose the configuration with the best validation Pearson correlation. The selected hyperparameters are reported in Table~\ref{tab:capt_backbone} and ~\ref{tab:capt_decoder}. For forecasting and decoding fine-tuning, we adopt the same fine-tuning strategy as described above for CalM. We train for 200 epochs during pretraining and all fine-tuning tasks. Checkpoints are selected based on validation Pearson correlation for neural forecasting and validation $R^2$ for behavior decoding.

\begin{table}[!htbp]
  \caption{Hyperparameters used for CAPT backbone}
  \label{tab:capt_backbone}
  \centering
  \begin{tabular}{ll}
    \toprule
    Hyperparameter & Value \\
    \midrule
    Patch length & 8 \\
    Model dimension & 512 \\
    Layers & 6 \\
    Attention heads & 8 \\
    FFN dimension & 2048 \\
    Dropout & 0.10 \\
    Embedding dropout & 0.10 \\
    Attention dropout & 0.10 \\
    Sampled neurons per batch & 512 \\
    \bottomrule
  \end{tabular}
\end{table}

\begin{table}[!htbp]
  \caption{Hyperparameters used for CAPT downstream decoder head}
  \label{tab:capt_decoder}
  \centering
  \begin{tabular}{ll}
    \toprule
    Hyperparameter & Value \\
    \midrule
    Decoder rank & 16 \\
    Head dropout & 0.6 \\
    Temporal smoothing kernel size & 9 \\
    \bottomrule
  \end{tabular}
\end{table}

\section{Ablation studies}
\label{app:ablation}

We study the effect of patch length via ablation experiments on a 12-session pretraining subset. As shown in the Table~\ref{patchlen-table}, increasing the patch length from 2 to 8 consistently improves forecasting performance. While a further increase to 10 yields comparable results, it does not lead to additional gains. Overall, these results suggest that a moderately large patch length is beneficial for model performance, and we adopt 8 as the default setting in all main experiments.

\begin{table}[!htbp]
  \caption{Ablation results for patch length.}
  \label{patchlen-table}
  \centering
  \begin{tabular}{ll}
    \toprule
    Patch Length & AR Corr \\
    \midrule
    2  & $0.2201 \pm 0.1073$ \\
    4  & $0.3689 \pm 0.0899$ \\
    8  & $0.4421 \pm 0.0680$ \\
    10 & $0.4406 \pm 0.0651$ \\
    \bottomrule
  \end{tabular}
\end{table}

In addition, we conduct ablation experiments on a 3-session subset to demonstrate the advantages of applying the residual learning, dynamic gating and QKV modulation for predictive performance and NeuroPAL classification under transferring setup in Table~\ref{tab:ablation-components} and~\ref{tab:neuropal-qkv-ablation}.

 \begin{table}[t]
\centering
\caption{Ablation study of CAPT components on the forecasting task. Results are reported as session-level mean $\pm$ standard deviation across three test sessions.}
\label{tab:ablation-components}
\begin{tabular}{lc}
\toprule
Component setting & AR Corr \\
\midrule
Baseline & 0.4916 $\pm$ 0.0553 \\
+ Dynamic gating & 0.5109 $\pm$ 0.0509 \\
+ QKV modulation & 0.4980 $\pm$ 0.0491 \\
+ Residual-last & 0.5026 $\pm$ 0.0595 \\
+ All components & \textbf{0.5166 $\pm$ 0.0618} \\
\bottomrule
\end{tabular}
\end{table}

\begin{table}[t]
\centering
\caption{Ablation study of QKV modulation on NeuroPAL cell identity classification after embedding-only transfer. Results are balanced accuracy (\%), reported as mean $\pm$ standard deviation across five folds.}
\label{tab:neuropal-qkv-ablation}
\begin{tabular}{lcc}
\toprule
Evaluation task & Baseline & + QKV modulation \\
\midrule
Atanas internal
& $56.81 \pm 4.19$ & $\mathbf{61.16 \pm 2.47}$ \\
Copper internal
& $65.32 \pm 4.27$ & $\mathbf{67.06 \pm 6.44}$ \\
Pooled
& $38.63 \pm 2.24$ & $\mathbf{43.96 \pm 1.69}$ \\
Atanas $\rightarrow$ Copper
& $27.03 \pm 0.99$ & $\mathbf{28.25 \pm 1.46}$ \\
Copper $\rightarrow$ Atanas
& $28.23 \pm 0.97$ & $\mathbf{32.56 \pm 1.22}$ \\
\bottomrule
\end{tabular}
\end{table}

\section{Additional results}
\label{app:result}

\textbf{Extra datasets}

\begin{figure}[!htbp]
  \centering
  \includegraphics[width=\linewidth]{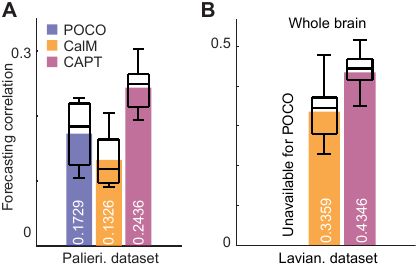}
  \caption{
  \textit{Performance of CAPT on additional transfer datasets.} (A) Neural population forecasting on Palieri. zebrafish datasets. (B) Neural population forecasting on the whole brain subset from Lavian. zebrafish datasets.
  }
  \label{fig:sup_fig_zebrafish}
\end{figure}

As additional evidence supporting the main forecasting results in Fig.~\ref{fig:fig3}, Fig.~\ref{fig:sup_fig_zebrafish} reports performance on two further zebrafish datasets: the Palieri dataset and the whole-brain recordings from Lavian et al. Across both datasets, CAPT continues to outperform the compared baselines. In particular, the whole-brain dataset contains pseudo-trials with a 16-frame context and a 16-frame prediction horizon. Pretrained POCO is not directly applicable for fine-tuning because its downstream forecasting setup requires the context and prediction horizon to remain consistent with the configuration used by the pretrained backbone. By contrast, the autoregressive designs of CalM and CAPT are naturally compatible with different context and rollout horizons, allowing them to adapt to this shorter pseudo-trial setting.

\textbf{Further investigation on POYO+ and POCO}

To further examine whether the reason fine-tuned POYO+ degrades on the \textit{C. elegans} datasets in Fig.~\ref{fig:fig3} is due to suboptimal hyperparameter choices, we train POYO+ directly on each target \textit{C. elegans} dataset using the same model configuration and hyperparameters as in the main experiments. As shown in Table~\ref{pretrain-table}, in-domain POYO+ training substantially improves decoding performance compared with cross-species fine-tuning, indicating that the degradation observed in Fig.~\ref{fig:fig3} is unlikely to be explained simply by inappropriate hyperparameters or insufficient model capacity.

\begin{table}[!htbp]
  \caption{Pretraining POYO+ results for \textit{C. elegans} decoding.}
  \label{pretrain-table}
  \centering
  \begin{tabular}{llll}
    \toprule
    Dataset (behavior) & POYO+ fine-tuning & POYO+ pretraining & CAPT fine-tuning \\
    \midrule
    Atanas (head angle) &
    $-0.0124 \pm 0.1349$ &
    $0.4968 \pm 0.1715$ &
    $0.5918 \pm 0.1522$ \\
    \midrule
    Copper (velocity) &
    $0.2773 \pm 0.2149$ &
    $0.6814 \pm 0.1131$ &
    $0.7503 \pm 0.0948$ \\
    \bottomrule
  \end{tabular}
\end{table}

\begin{figure}[!htbp]
  \centering
  \includegraphics[width=\linewidth]{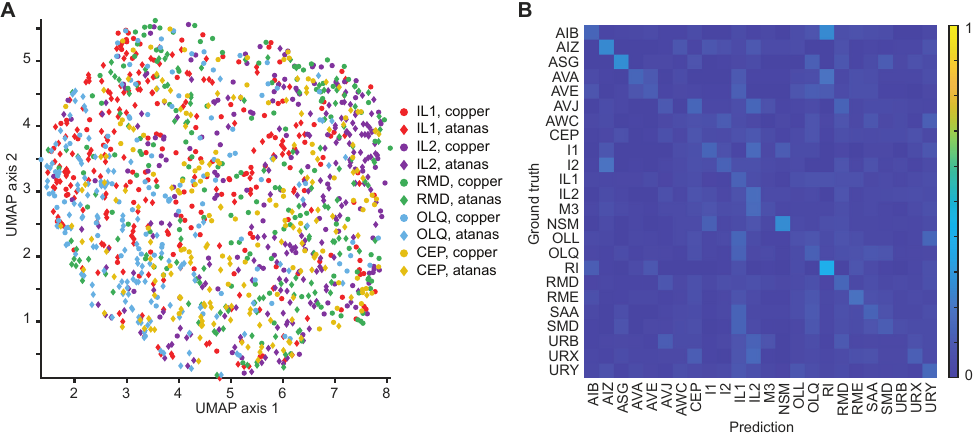}
  \caption{
  \textit{Embedding analysis for POCO baseline.} (A) 2d UMAP visualization for POCO neural embeddings (B) Confusion matrix of cell-type classification from POCO neural embeddings
  }
  \label{fig:sup_fig_poco}
\end{figure}

As additional evidence supporting the main embedding analysis results in Fig~\ref{fig:fig4}, Fig~\ref{fig:sup_fig_poco} shows that neural embeddings learned by POCO exhibit less structured organization in the UMAP visualization and weaker cell-type separability in the corresponding confusion matrix. Compared with CAPT embeddings, POCO embeddings show less coherent clustering by NeuroPAL-defined cell-type, and the classifier produces a less accurate and more diffuse confusion pattern.

\textbf{Statistical test}

We report the mean ± standard deviation and exact p-values in the experiments in Table~\ref{tab:statistical_test}. All methods are evaluated on identical session-level splits and compared using two-sided paired Wilcoxon signed-rank tests. CAPT significantly outperforms POCO/POYO+ in 17 of 20 comparisons and CalM in 16 of 20; the nonsignificant cases involve three-session datasets, where the paired Wilcoxon test cannot reach $p<0.25$, or closely matched performance.

\textbf{Trace-statistic control}

To examine whether the NeuroPAL classification performance can be explained by
simple calcium trace properties, we construct handcrafted trace-statistic
features and evaluate them using the same linear SVM classifier and validation
protocol as the embedding analysis. For each neuron, we extract a 45-dimensional
feature vector from the full calcium trace, including signal statistics,
first-order variability, temporal autocorrelation, autocorrelation decay,
variance decomposition, Hjorth parameters, noise and SNR estimates, and
frequency-domain characteristics. These features are computed without using
cell-identity labels.

Table~\ref{tab:neuropal_trace_control} reports the balanced accuracy. Trace
statistics achieve above-chance performance only in the pooled setting, while
cross-dataset transfer remains close to chance, suggesting that simple signal
properties alone are insufficient to explain the transferable cell-identity
information captured by CAPT embeddings.

\begin{table}[t]
\centering
\caption{NeuroPAL classification using handcrafted calcium trace statistics.}
\label{tab:neuropal_trace_control}
\begin{tabular}{lc}
\toprule
Setup & Balanced Accuracy \\
\midrule
Pooled & 0.1663 $\pm$ 0.0107 \\
Atanas $\rightarrow$ Copper & 0.0479 $\pm$ 0.0060 \\
Copper $\rightarrow$ Atanas & 0.0722 $\pm$ 0.0157 \\
\bottomrule
\end{tabular}
\end{table}

\begin{table}[t]
\centering
\caption{Comparison of CAPT with POCO / POYO+ and CalM across forecasting and decoding tasks. Values are reported as mean $\pm$ standard deviation.}
\label{tab:statistical_test}
\resizebox{\linewidth}{!}{
\begin{tabular}{llllll}
\toprule
Result & POCO / POYO+ & CalM & CAPT & $p$ (vs. POCO / POYO+) & $p$ (vs. CalM) \\
\midrule
Tseng held-in forecasting corr ($n=189$)
& $0.3668 \pm 0.0605$
& $0.3799 \pm 0.0978$
& $\mathbf{0.5155 \pm 0.0655}$
& $9.136\times10^{-33}$ (***)
& $9.136\times10^{-33}$ (***) \\
Tseng held-out forecasting corr ($n=97$)
& $0.3322 \pm 0.0511$
& $0.3383 \pm 0.0860$
& $\mathbf{0.4677 \pm 0.0556}$
& $1.218\times10^{-17}$ (***)
& $1.218\times10^{-17}$ (***) \\
Sun forecasting corr ($n=52$)
& $0.2553 \pm 0.0378$
& $0.2608 \pm 0.0660$
& $\mathbf{0.3353 \pm 0.0578}$
& $3.504\times10^{-10}$ (***)
& $3.504\times10^{-10}$ (***) \\
Brysch forecasting corr ($n=29$)
& $0.1872 \pm 0.0239$
& $0.1342 \pm 0.0248$
& $\mathbf{0.2129 \pm 0.0249}$
& $4.098\times10^{-7}$ (***)
& $3.725\times10^{-9}$ (***) \\
Copper forecasting corr ($n=27$)
& $0.2161 \pm 0.1070$
& $0.2075 \pm 0.1142$
& $\mathbf{0.2823 \pm 0.1298}$
& $1.843\times10^{-4}$ (***)
& $7.290\times10^{-5}$ (***) \\
CaLiAli forecasting corr ($n=3$)
& $0.2296 \pm 0.0076$
& $0.1119 \pm 0.0282$
& $\mathbf{0.2970 \pm 0.0082}$
& $0.2500$ (n.s.)
& $0.2500$ (n.s.) \\
Park forecasting corr ($n=3$)
& $0.1908 \pm 0.0141$
& $0.1333 \pm 0.0228$
& $\mathbf{0.2576 \pm 0.0173}$
& $0.2500$ (n.s.)
& $0.2500$ (n.s.) \\
Landmark forecasting corr ($n=27$)
& $0.4097 \pm 0.0694$
& $0.1968 \pm 0.0979$
& $\mathbf{0.4805 \pm 0.0576}$
& $1.490\times10^{-8}$ (***)
& $1.490\times10^{-8}$ (***) \\
aHB forecasting corr ($n=3$)
& $0.3289 \pm 0.0172$
& $0.0952 \pm 0.0324$
& $\mathbf{0.4003 \pm 0.0197}$
& $0.2500$ (n.s.)
& $0.2500$ (n.s.) \\
Atanas forecasting corr ($n=80$)
& $0.2058 \pm 0.1075$
& $0.1780 \pm 0.0846$
& $\mathbf{0.2575 \pm 0.1062}$
& $3.992\times10^{-10}$ (***)
& $3.619\times10^{-11}$ (***) \\
\midrule
Tseng held-in decoding pitch $R^2$ ($n=189$)
& $0.7227 \pm 0.0840$
& $0.7873 \pm 0.0737$
& $\mathbf{0.8458 \pm 0.0623}$
& $9.135\times10^{-33}$ (***)
& $9.136\times10^{-33}$ (***) \\
Tseng held-in decoding roll $R^2$ ($n=189$)
& $0.6751 \pm 0.1132$
& $0.7540 \pm 0.0883$
& $\mathbf{0.8050 \pm 0.0759}$
& $9.135\times10^{-33}$ (***)
& $1.054\times10^{-32}$ (***) \\
Tseng held-in decoding yaw $R^2$ ($n=189$)
& $0.5853 \pm 0.2321$
& $0.7076 \pm 0.1691$
& $\mathbf{0.7314 \pm 0.1649}$
& $2.076\times10^{-30}$ (***)
& $4.948\times10^{-17}$ (***) \\
Tseng held-out decoding pitch $R^2$ ($n=97$)
& $0.7073 \pm 0.0611$
& $0.7309 \pm 0.0785$
& $\mathbf{0.8114 \pm 0.0656}$
& $1.296\times10^{-17}$ (***)
& $1.218\times10^{-17}$ (***) \\
Tseng held-out decoding roll $R^2$ ($n=97$)
& $0.6446 \pm 0.1299$
& $0.6980 \pm 0.1200$
& $\mathbf{0.7739 \pm 0.0897}$
& $1.218\times10^{-17}$ (***)
& $1.218\times10^{-17}$ (***) \\
Tseng held-out decoding yaw $R^2$ ($n=97$)
& $0.5430 \pm 0.2240$
& $0.6022 \pm 0.2172$
& $\mathbf{0.6385 \pm 0.2348}$
& $5.331\times10^{-11}$ (***)
& $4.636\times10^{-9}$ (***) \\
Sun decoding $R^2$ ($n=52$)
& $0.8586 \pm 0.1438$
& $0.9521 \pm 0.0658$
& $\mathbf{0.9574 \pm 0.0466}$
& $3.504\times10^{-10}$ (***)
& $0.4335$ (n.s.) \\
Brysch decoding $R^2$ ($n=29$)
& $0.2796 \pm 0.3976$
& $0.3673 \pm 0.1696$
& $\mathbf{0.5250 \pm 0.1503}$
& $1.144\times10^{-6}$ (***)
& $1.193\times10^{-5}$ (***) \\
Copper decoding velocity $R^2$ ($n=27$)
& $0.2773 \pm 0.2149$
& $0.5882 \pm 0.2213$
& $\mathbf{0.7503 \pm 0.0948}$
& $1.490\times10^{-8}$ (***)
& $1.879\times10^{-5}$ (***) \\
Atanas decoding head angle $R^2$ ($n=80$)
& $-0.0124 \pm 0.1349$
& $0.0896 \pm 0.2729$
& $\mathbf{0.5918 \pm 0.1522}$
& $7.850\times10^{-15}$ (***)
& $7.850\times10^{-15}$ (***) \\
\bottomrule
\end{tabular}
}
\end{table}

\begin{table}[t]
\centering
\caption{Effect of adaptation into the frozen pretrained backbone on
cross-dataset NeuroPAL cell-identity classification. Embedding-only adaptation
updates target neuron and session embeddings against the frozen mouse-pretrained backbone, while independent retraining trains CAPT from scratch on each \textit{C. elegans} dataset separately. Results are mean $\pm$ standard
deviation across five folds; chance level is $0.0435$.}
\label{tab:neuropal_retrain}
\begin{tabular}{llcc}
\toprule
Adaptation & Direction & Balanced accuracy & Macro F1 \\
\midrule
Embedding-only (frozen backbone) & Atanas $\rightarrow$ Copper & $\mathbf{0.3645 \pm 0.0091}$ & $\mathbf{0.3814 \pm 0.0084}$ \\
Embedding-only (frozen backbone) & Copper $\rightarrow$ Atanas & $\mathbf{0.3535 \pm 0.0079}$ & $\mathbf{0.3723 \pm 0.0084}$ \\
Independent retraining           & Atanas $\rightarrow$ Copper & $0.0373 \pm 0.0034$ & $0.0199 \pm 0.0025$ \\
Independent retraining           & Copper $\rightarrow$ Atanas & $0.0388 \pm 0.0020$ & $0.0152 \pm 0.0037$ \\
\bottomrule
\end{tabular}
\end{table}

\textbf{Role of adaptation into the frozen backbone}

The cross-dataset results in Fig.~\ref{fig:fig4}D compare embeddings that were
adapted into a single frozen pretrained backbone. To test whether this shared
coordinate system, rather than the architecture or the per-neuron embedding
capacity, is what makes embeddings comparable across datasets, we retrain CAPT
from scratch independently on each \textit{C. elegans} dataset and repeat the
identical cross-dataset classification protocol. As shown in
Table~\ref{tab:neuropal_retrain}, cross-dataset transfer collapses to chance in
both directions, even though each model is trained on the same traces with the
same architecture and the same number of embedding parameters. Independently
learned embedding spaces are therefore not mutually comparable, indicating that
adaptation into a shared frozen backbone, and not the embedding parameterization
itself, is what renders neuron embeddings commensurable across datasets.

\section{Computational resource}
\label{app:compute}
Experiments were conducted using one GPU node equipped with 8× NVIDIA A100 GPUs (40 GB memory each). Models were trained using distributed data parallelism across available GPUs.



\end{document}